\documentclass[10pt,twocolumn,letterpaper]{article}

\usepackage[pagenumbers]{cvpr}     
\usepackage{graphicx}
\usepackage{amsmath}
\usepackage{amssymb}
\usepackage{booktabs}

\usepackage{algorithm}
\usepackage{algorithmic}
\usepackage{multirow}
\usepackage{bbding}

\usepackage{enumitem}

\usepackage{color, colortbl}
\usepackage[dvipsnames]{xcolor}
\usepackage[pagebackref,breaklinks,colorlinks]{hyperref}

\usepackage{multirow}

\newcommand{\tableCellHeight}{1}
\newcommand{\tabstyle}[1]{
  \setlength{\tabcolsep}{#1}
  \renewcommand{\arraystretch}{\tableCellHeight}
  \centering
  \small
}

\newcommand{\hgreen}[1]{\textcolor{ForestGreen}{#1}} 
\definecolor{tabhighlight}{HTML}{e5e5e5}

\usepackage[capitalize]{cleveref}
\crefname{section}{Sec.}{Secs.}
\Crefname{section}{Section}{Sections}
\Crefname{table}{Table}{Tables}
\crefname{table}{Tab.}{Tabs.}

\def\ie{\emph{i.e}\onedot}

\newcommand*\samethanks[1][\value{footnote}]{\footnotemark[#1]}

\begin{document}


\title{Task-Oriented Multi-Modal Mutual Leaning for Vision-Language Models}

\author{Sifan Long\textsuperscript{\rm 1,2}
\thanks{Equal contribution.}
~~~~Zhen Zhao\textsuperscript{\rm 3,2}\samethanks~~~~Junkun Yuan\textsuperscript{\rm 4,2}\samethanks
~~~~Zichang Tan\textsuperscript{\rm 2}
~~~~Jiangjiang Liu\textsuperscript{\rm 2} \\
~~~~Luping Zhou\textsuperscript{\rm 3}
~~~~Shengsheng Wang\textsuperscript{\rm 1}\thanks{Corresponding authors.}
~~~~Jingdong Wang\textsuperscript{\rm 2}\samethanks \\
\textsuperscript{\rm 1}Jilin University\hspace{8mm}
\textsuperscript{\rm 2}Baidu VIS\hspace{8mm}
\textsuperscript{\rm 3}University of Sydney\hspace{8mm}
\textsuperscript{\rm 4}Zhejiang University\\
{\tt\small longsf22@mails.jlu.edu.cn~~\{zhen.zhao, luping.zhou\}@sydney.edu.au
~~yuanjk@zju.edu.cn}\\
{\tt\small wss@jlu.edu.cn~~\{tanzichang, liujiangjiang, wangjingdong\}@baidu.com}
}

\maketitle 
\begin{abstract}
Prompt learning has become one of the most efficient paradigms for adapting large pre-trained vision-language models to downstream tasks. 
Current state-of-the-art methods, like CoOp and ProDA, tend to adopt soft prompts to learn an appropriate prompt for each specific task. Recent CoCoOp further boosts the base-to-new generalization performance via an image-conditional prompt. 
However, it directly fuses identical image semantics to prompts of different labels and significantly weakens the discrimination among different classes as shown in our experiments.
Motivated by this observation, we first propose a class-aware text prompt (CTP) to enrich generated prompts with label-related image information. Unlike CoCoOp, CTP can effectively involve image semantics and avoid introducing extra ambiguities into different prompts.
On the other hand, instead of reserving the complete image representations, we propose text-guided feature tuning (TFT) to make the image branch attend to class-related representation.
A contrastive loss is employed to align such augmented text and image representations on downstream tasks.
In this way,  the \textbf{image-to-text}  CTP and \textbf{text-to-image} TFT can be mutually promoted to enhance the adaptation of VLMs for downstream tasks.
Extensive experiments demonstrate that our method outperforms the existing methods by a significant margin. Especially, compared to CoCoOp, we achieve an average improvement of 4.03\% on new classes and 3.19\% on harmonic-mean over eleven classification benchmarks.

\end{abstract}

\begin{figure}
	\centering
	\includegraphics[width=0.49\textwidth]{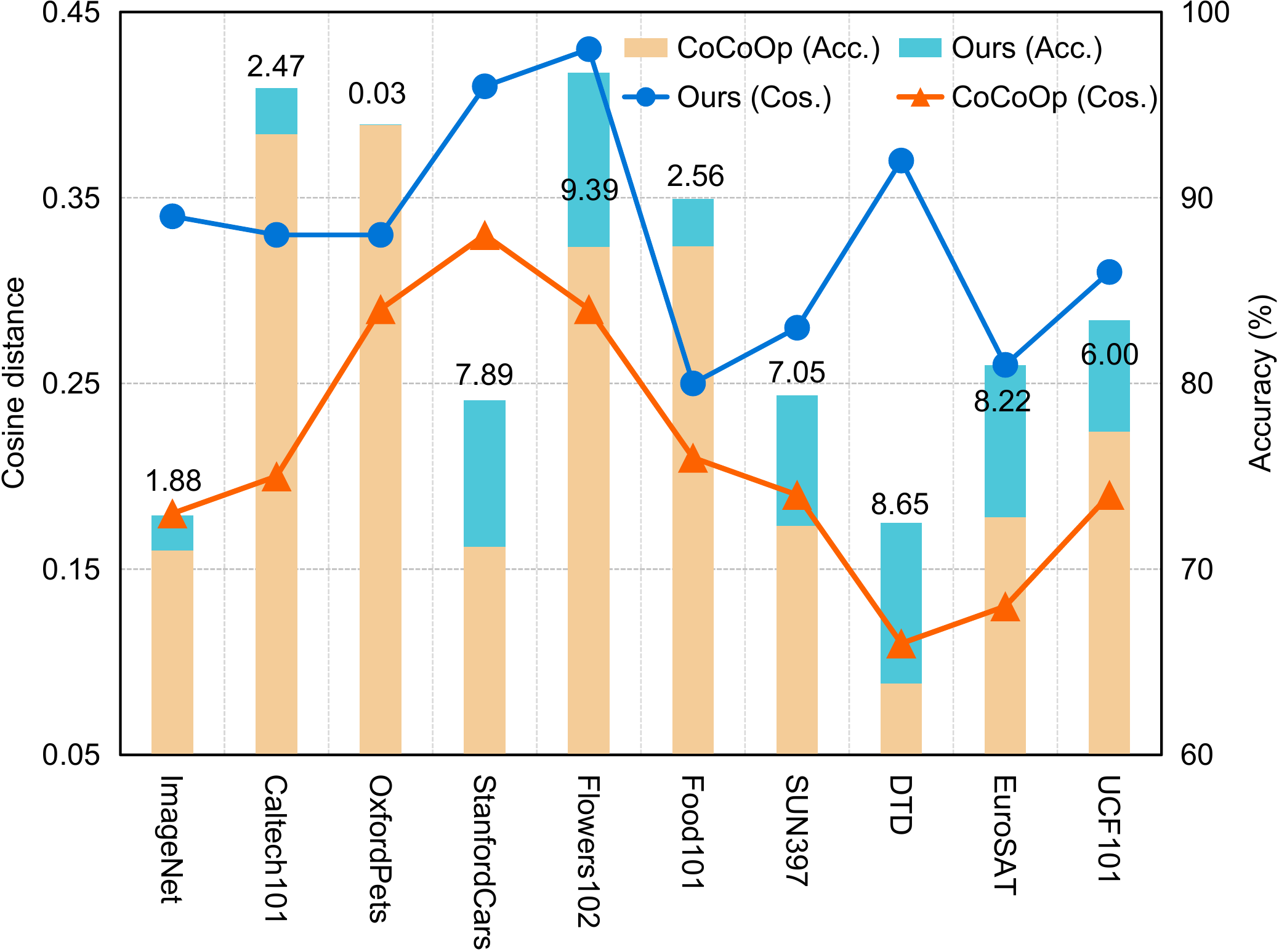}
	\caption{Comparisons between CoCoOp~\cite{zhou2022conditional} and our method.
    The cosine distance between 
    the positive and the negative prompts, 
    which quantifies the class discrimination, and the average accuracy on  benchmarks are reported.
	}
	\label{FIG:1}
\end{figure}

\section{Introduction}
\label{sec:intro}

Recently, large vision-language models (VLM), such as CLIP \cite{radford2021learning} and ALIGN \cite{jia2021scaling}, which employ language as supervision signal instead of discrete labels, have shown impressive generalization performance in a wide range of downstream vision tasks. Their multi-modal interaction nature delivers open-vocabulary support and achieves amazing zero-shot classification performance. Despite their impressive transferable abilities, as discussed in \cite{liu2023pre}, it is essential to re-activate specific representation capabilities for optimal performance in certain downstream tasks.
Considering their hundreds of millions or billions of parameters, attempting to fine-tune the entire model is impractical and even jeopardizes the well-established representation space \cite{houlsby2019parameter}. 
To this end, many recent studies have centred on the efficient and effective adaptation of pre-trained and frozen large VLMs for the specific downstream tasks \cite{lu2022prompt,zhou2022learning, zhou2022conditional}.

Prompt, a simple, compact, and viable strategy, has become the leading solution for deploying large pre-trained VLMs into certain downstream tasks.
CLIP \cite{radford2021learning} utilizes hand-crafted prompts to achieve impressive zero- and few-shot classification performance. Nevertheless, manually-designed prompts require significant domain knowledge and can be highly time-consuming and sub-optimal for specific downstream tasks.
To address this problem, later studies \cite{lu2022prompt,zhou2022learning} adopt soft prompts to learn an appropriate text prompt via optimizing a contrastive loss on different text labels.
CoCoOp~\cite{zhou2022conditional} further highlights the limitations of such static soft prompts and proposes learning \textbf{image-dependent} prompts conditioned on individual instances rather than fixed prompts. It achieves great performance gains on unseen classes by adding high-level image embedding to text prompts.
However, compared to CoOp with static prompts, CoCoOp essentially fuses identical image semantics with different text labels, leading to inevitable learning ambiguity and resulting in an average performance drop of 2.22\% on base classes on 11 datasets (see Table \ref{tab:results_generalization}). For example, it may associate the dog image semantics with a prompt that references the [class] of a cat. 
When using the cosine distance to measure the differences between the positive and negative text prompts, as shown in \cref{FIG:1}, CoCoOp holds low distance values, suggesting that it brings significant learning ambiguities to text prompts.
Therefore, we argue that text prompts should not only condition on distinct input images for better generalization abilities, but also adapt to different classes to eliminate the potential ambiguities.

To achieve this goal, we propose Class-aware Text Prompts (CTP), which leverages label-related image information to generate finer prompts. 
Specifically, we first contact learnable context vectors and each class label to model the initial prompt sentences. Then we leverage these class prompt sentences to query their corresponding image regions and representations. Corresponding related image features are subsequently added to initial class prompt sentences to produce the final text prompts. In this way, generated image-dependent and class-aware prompts can better concentrate on the image information in a more precise manner. As shown in \cref{FIG:1}, our method enjoys better discrimination between positive and negative prompts and consistently outperforms CoCoOp on 11 classification datasets.

\begin{figure}
	\centering
	\includegraphics[width=0.45\textwidth]{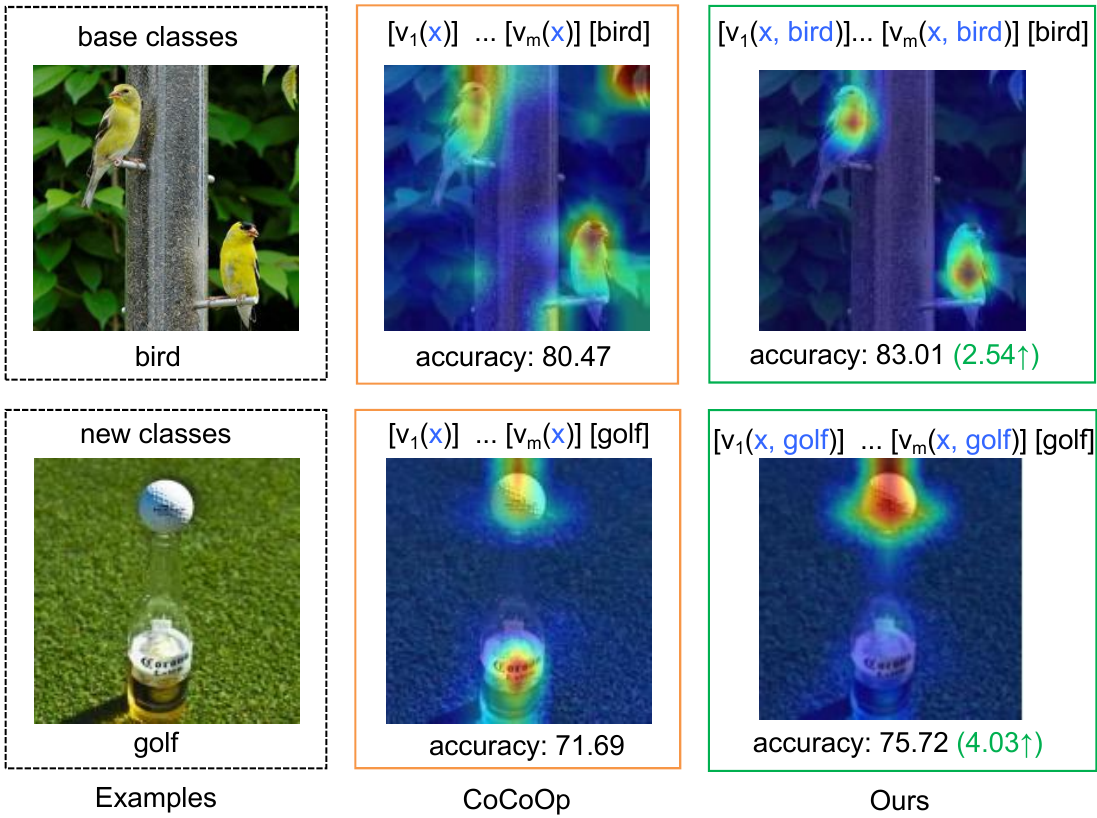}
	\caption{Comparisons of attention map visualization for CoCoOp and our method on ImageNet. 
    Our method obtains better average accuracy of both base and new classes across 11 datasets by paying attention to task-related regions.
	}
	\label{FIG:2}
\end{figure}

On the other hand, we identify a critical problem in these text prompt-based strategies: the image branch is ignored and not adjusted to specific downstream tasks.
As shown in \cref{FIG:2} (CoCoOp), on the task of identifying birds, the output image feature, without further tuning, can be distracted to leaves of the same color. Similarly, it also wrongly highlights the beer foam that is of a similar shape to recognize golf balls.
Since the final recognition is jointly inferred by both text and image branches, such an issue may degrade the classification performance.
Thus it is necessary to tune the image features further so that the image branch can focus more on the tasks-related representation.
We then propose Text-guided Feature Tuning (TFT), which leverages encoded text embedding to guide image representation more on task-related regions. 
As shown in \cref{FIG:2} (ours), our method successfully focuses on task-related regions, \ie, birds and golf balls. 
We then leverage the contrastive loss function to further align class-aware text embedding and text-guided image features on certain downstream tasks.

In summary, we propose a new task-oriented multi-modal mutual-learning method, which well-integrates our designed class-aware text prompts and text-guided feature tuning for fast adaptation of frozen VLMs on downstream tasks. Image features can help construct image-dependant class-aware text prompts, leading to more discriminative text embedding. Simultaneously, improved text embedding can further guide the image branch attending to class-related representation. In this way, these two different modality branches can be tightly coupled and mutual-beneficial 
across the whole training process. 
Our main contributions are summarized in the following.
\begin{itemize}
    \item We propose class-aware text prompts which generate prompts based on task-relevant image semantics instead of complete visual information. In this way, we improve the classification accuracy of unseen classes without introducing extra learning ambiguities.
    \item We propose text-guided feature tuning which enforces image branch to pay more attention to the task-related representation. 
    As a result, the model avoids deviating attention to the task-irrelevant regions of the image.
    \item Benefiting from our mutual learning strategy, our method achieves SOTA results on four downstream tasks. Especially, ours significantly outperforms existing methods on the base-to-new generalization task.
\end{itemize}

\begin{figure*}[t]
	\centering
	\includegraphics[width=1.\textwidth]{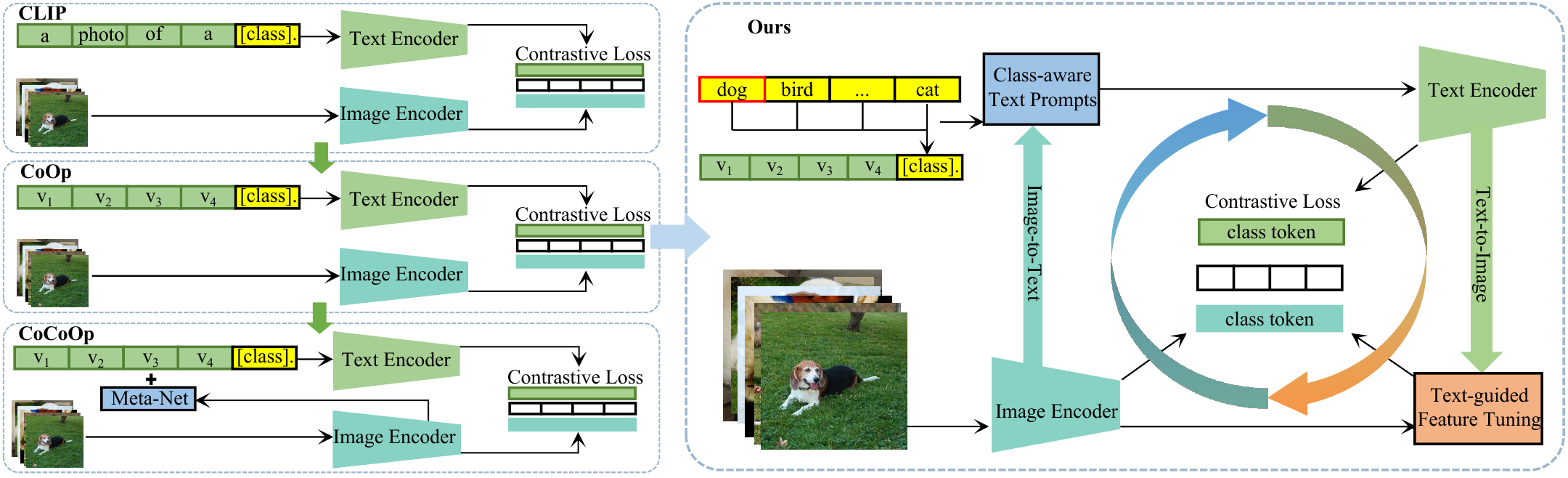}
	\caption{ 
    Comparisons of three representative prompt learning techniques and our method.
    The main differences lie in how the text and image branches focus on downstream tasks. CLIP artificially designs prompt templates. CoOp designs automatic prompts using learnable parameters. CoCoOp directly allows text branch to focus on images semantic through Meta-Net. 
    We introduce Class-aware Text Prompts (CTP) and Text Feature Tuning (TFT) to the text and image branches, respectively.
    The CTP generates prompts based on class-related image information instead of using the identical image semantics like CoCoOp. The TFT enables the image branch to directly focus on downstream tasks. We leverage the contrastive loss function to align task-oriented text and images, making them promote each other for achieving better downstream generalization performance.
	}
	\label{FIG:3}
 \vspace{-0.2cm}
\end{figure*}
\section{Related work}
\label{sec:rwork}
\paragraph{Vision language models (VLM).} 
The current VLM can be roughly divided into four categories based on the training objectives: image-text matching \cite{chen2020uniter, li2019visualbert, lu2019vilbert}, contrastive loss \cite{li2021align, li2020unimo, li2020oscar}, masked language modeling \cite{su2019vl, tan2019lxmert, yu2021ernie}, and masked image modeling \cite{chen2020uniter, lu2019vilbert,su2019vl}. 
As a milestone,  CLIP utilizes 400 million image-text pairs to train a large-scale multi-modal model and demonstrates promising performance on a wide spectrum of tasks including few-shot and zero-shot visual recognition.
Motivated  by this work, numerous follow-ups have been proposed  to improve the effectiveness (e.g., FLIP \cite{li2022scaling}, A-CLIP \cite{yang2022attentive}, MaskCLIP \cite{dong2022maskclip}, and SLIP \cite{mu2022slip}) or apply it to other domains (e.g., DenseCLIP \cite{rao2022denseclip} and ActionCLIP \cite{wang2021actionclip}). 
The primary limitation of these methods is that hand-crafted prompts are dataset-sensitive and difficult to optimize. We design an automatic and learnable prompts method to enhance the generalization performance of pre-trained models on downstream tasks.
\vspace{-0.2cm}
\paragraph{Prompt learning in NLP.} 
As the scale and complexity of pre-trained language models continue to grow, fine-tuning for specific tasks is becoming increasingly expensive. 
In contrast, prompt-based approaches are an efficient and lightweight alternative that can be used to generate high-quality text with much lower computational requirements.
The original prompts were manually designed prompt templates. 
While manually designing prompts is advantageous due to their intuitive and comprehensible nature, it also presents a significant challenge that demands extensive experimentation, experience, and language expertise, resulting in high costs.
To overcome the limitations of manual prompt design, numerous studies have initiated research into automatically learn appropriate prompts. 
The automatic prompts can be categorized into two types: discrete prompts and continuous prompts. Discrete prompts consist of various approaches such as prompt mining \cite{jiang2020can}, prompt paraphrasing \cite{yuan2021bartscore,haviv2021bertese}, gradient-based search \cite{wallace2019universal}, prompt generation \cite{gao2020making} and prompt scoring \cite{davison2019commonsense}. On the other hand, continuous prompts include techniques such as prefix tuning \cite{li2021prefix}, tuning initialized with discrete prompts \cite{shin2020autoprompt} and hard-soft prompt hybrid tuning \cite{liu2021gpt}. 
These methods have also been applied to the field of computer vision for prompt learning research.
However, the task of prompt learning in computer vision is often considered more challenging than in natural language due to the relatively limited high-level semantic information present in visual data with raw pixels.
\vspace{-0.6cm}

\paragraph{Prompt learning in vision language models.} 
Prompt learning has been demonstrated to be an effective method for improving the performance of pre-trained language models on downstream tasks.  Recently, prompt learning has gained increased attention in the context of vision language models.
For example, CoOp \cite{zhou2022learning} employs learnable vectors to model contextual words as prompts,  and demonstrates that automatic prompts outperform hand-crafted prompts in downstream tasks. 
CoCoOp \cite{zhou2022conditional} extends CoOp by incorporating lightweight neural networks to dynamically generate prompts based on each image, thus mitigating sensitivity to class shifts. 
Different from the above methods, VP \cite{bahng2022exploring}, VPT \cite{wu2022unleashing}, and EVP \cite{jia2022visual} prompt with images.
VP \cite{bahng2022exploring} directly combines learnable prompts and pixel-wise input images as new inputs to the model.
EVP \cite{wu2022unleashing} shrinks the original image before padding the prompts around it, to avoid destroying the original image information. 
VPT \cite{jia2022visual} introduces a small amount learnable parameters into the input sequence of each transformer layer and learns them together with a linear head during fine-tuning. 
Building on the prompt learning approach of the text branch, we propose class-aware text prompt that generates image-dependent and class-aware prompts. Similarly, follow the feature tuning of image branch, we introduce text-guided tuning, which directs the image branch to focus on the task-relevant local regions rather than the global information.


\section{Method}
\label{sec:method}
\subsection{Comparisons of CLIP, CoOp, and CoCoOp}
\paragraph{CLIP} comprises two encoders: an image encoder
and a text encoder. The image encoder, denotes by $F(x)$, converts an image $x \in \mathbb{R}^{3 \times H \times W}$ with height of $H$ and width of $W$ into a $d$-dimensional image feature ${f}_x  \in \mathbb{R}^{N \times d}$, where $N$ is the number of split patches. Meanwhile, the text encoder, denoted as $G(t)$, generates an $d$-dimensional text representation ${g}_t  \in \mathbb{R}^{M \times d} $ from natural language text $t$, where $M$ is the number of classes.
Two encoders are jointly trained using a contrastive loss function that maximizes the cosine similarity of matched pairs and minimizes that of the unmatched pairs.
After training, CLIP can be directly used for zero-shot image recognition without requiring fine-tuning of the whole model. Since CLIP is pre-trained on whether an image matches a textual description, the hand-crafted prompt template is employed to convert raw labels into textual descriptions. The most common form of template in CLIP is ``a photo of a [CLASS]", where the class token is replaced with specific class names such as ``cat", ``dog", ``car", etc. We let the image features ${f}_x$ of an image $x$ be extracted by an image encoder and the text features ${g}_t$ be obtained by feeding the prompt description into the text encoder. 
The prediction task is defined as the classification of an image into one of $C$ categories, which are represented by the set $y \in\{1, \ldots, C\}$. Denote $y$ as the predicted category.
Let $g_t^i$ be the $i$-th dimension of text features $g_t$, with image features $f_x$, we have the predicted probability of the $i$-th class:
\begin{equation}
P(y=i \mid x)=\frac{\exp \left(\cos \left({f}_x, {g}^i_t\right) / \tau\right)}{\sum_{j=1}^C \exp \left(\cos \left({f}_x, {g}^j_t\right) / \tau\right)},
\end{equation}
where $\cos (\cdot, \cdot)$ denotes the cosine similarity and $\tau$ is the temperature parameter of the softmax function.

\begin{table*}[t]
    \tabstyle{6pt}
    \setlength{\tabcolsep}{1.mm}{
    \begin{subtable}[t]{.3\textwidth}
    \centering
    \caption{\textbf{Average over 11 datasets}}
    \begin{tabular}{l cc|c}
    \toprule
    & Base & New & Hos \\
    \midrule
    CLIP     & 69.34 & 74.22 & 71.70 \\
    CoOp    & 82.69 & 63.22 & 71.66 \\
    CoCoOp  & 80.47 & 71.69 & 75.83\\
    ProDA   & 81.56 & 72.30 & 76.65\\
    \rowcolor{tabhighlight}
    Ours &\bf 83.01   & \textbf{75.72}	 & \textbf{79.02}\\ 
    \bottomrule
    \end{tabular}
    \end{subtable}
    \vspace{1em}
    \begin{subtable}[t]{.3\textwidth}
    \centering
    \caption{ImageNet}
    \begin{tabular}{l cc|c}
    \toprule
    & Base & New & Hos \\
    \midrule
    CLIP   & 72.43 & 68.14 & 70.22\\
    CoOp   & 76.47 & 67.88 & 71.92\\
    CoCoOp & 75.98 & 70.43 & 73.10\\
    ProDA  & 75.40 & 70.23 & 72.72\\
    \rowcolor{tabhighlight}
    Ours & \textbf{77.42} & \bf 70.44 & \textbf{73.77}\\
    \bottomrule
    \end{tabular}
    \end{subtable}
    ~
    \begin{subtable}[t]{.3\textwidth}
    \centering
    \caption{Caltech101}
    \begin{tabular}{l cc|c}
    \toprule
    & Base & New & Hos \\
    \midrule
    CLIP   & 96.84 & 94.00 & 95.40 \\
    CoOp   & 98.00 & 89.81 & 93.73 \\
    CoCoOp & 97.96 & 93.81 & 95.84 \\
    ProDA  & 98.27 & 93.23 & 95.68\\
    \rowcolor{tabhighlight}
    Ours & \textbf{98.31} & \bf94.75 & \textbf{96.50}\\
    \bottomrule
    \end{tabular}
    \end{subtable}
    ~
    \begin{subtable}[t]{.3\textwidth}
    \centering
    \caption{OxfordPets}
    \begin{tabular}{l cc|c}
    \toprule
    & Base & New & Hos \\
    \midrule
    CLIP    & 91.17 & 97.26 & 94.12 \\
    CoOp    & 93.67 & 95.29 & 94.47 \\
    CoCoOp  & 95.20 & 97.69 & 96.43 \\
    ProDA   & 95.43 & \bf 97.83 & 96.62 \\
    \rowcolor{tabhighlight}
    Ours & \bf 95.86 & 97.55 & \bf96.70\\
    \bottomrule
    \end{tabular}
    \end{subtable}
    \vspace{1em}
    \begin{subtable}[t]{.3\textwidth}
    \centering
    \caption{StanfordCars}
    \begin{tabular}{l cc|c}
    \toprule
    & Base & New & Hos \\
    \midrule
    CLIP   & 63.37 & \textbf{74.89} & 68.65 \\
    CoOp   & \textbf{78.12} & 60.40 & 68.13 \\
    CoCoOp & 70.49 & 73.59 & 72.01 \\
    ProDA  & 74.70 & 71.20 & 72.91\\
    \rowcolor{tabhighlight}
    Ours & 76.29 & 74.17 & \textbf{75.22} \\
    \bottomrule
    \end{tabular}
    \end{subtable}
    ~
    \begin{subtable}[t]{.3\textwidth}
    \centering
    \caption{Flowers102}
    \begin{tabular}{l cc|c}
    \toprule
    & Base & New & Hos \\
    \midrule
    CLIP   & 72.08 & 77.80 & 74.83 \\
    CoOp   & 97.60 & 59.67 & 74.06 \\
    CoCoOp & 94.87 & 71.75 & 81.71 \\
    ProDA  & \bf 97.70 & 68.68 & 80.66 \\
    \rowcolor{tabhighlight}
    Ours &  97.36 & \textbf{77.70} & \textbf{86.43} \\
    \bottomrule
    \end{tabular}
    \end{subtable}
    ~
    \begin{subtable}[t]{.3\textwidth}
    \centering
    \caption{Food101}
    \begin{tabular}{l cc|c}
    \toprule
    & Base & New & Hos \\
    \midrule
    CLIP   & 90.10 & 91.22 & 90.66 \\
    CoOp   & 88.33 & 82.26 & 85.19 \\
    CoCoOp & \bf 90.70 & 91.29 & 90.99 \\
    ProDA  & 90.30 & 88.57 & 89.43  \\
    \rowcolor{tabhighlight}
        Ours & 90.54 & \textbf{92.31} & \bf 91.42\\
    \bottomrule
    \end{tabular}
    \end{subtable}
    \vspace{1em}
    \begin{subtable}[t]{.3\textwidth}
    \centering
    \caption{FGVCAircraft}
    \begin{tabular}{l cc|c}
    \toprule
    & Base & New & Hos \\
    \midrule
    CLIP   & 27.19 & \textbf{36.29} & 31.09\\
    CoOp   & \bf40.44 & 22.30 & 28.75 \\
    CoCoOp & 33.41 & 23.71 & 27.74 \\
    ProDA  & 36.90 & 34.13 & 35.46 \\
    \rowcolor{tabhighlight}
    Ours & 39.49 & 35.37 & \textbf{37.32}\\
    \bottomrule
    \end{tabular}
    \end{subtable}
    ~
    \begin{subtable}[t]{.3\textwidth}
    \centering
    \caption{SUN397}
    \begin{tabular}{l cc|c}
    \toprule
    & Base & New & Hos \\
    \midrule
    CLIP   & 69.36 & 75.35 & 72.23 \\
    CoOp   & 80.60 & 65.89 & 72.51 \\
    CoCoOp & 79.74 & 76.86 & 78.27 \\
    ProDA  & 78.67 & 76.93 & 77.79 \\
    \rowcolor{tabhighlight}
    Ours & \textbf{82.16} & \bf 77.49 & \textbf{79.76}\\
    \bottomrule
    \end{tabular}
    \end{subtable}
    ~
    \begin{subtable}[t]{.3\textwidth}
    \centering
    \caption{DTD}
    \begin{tabular}{l cc|c}
    \toprule
    & Base & New & Hos \\
    \midrule
    CLIP   & 53.24 & 59.90 & 56.37 \\
    CoOp   & 79.44 & 41.18 & 54.24 \\
    CoCoOp & 77.01 & 56.00 & 64.85 \\
    ProDA  & \bf80.67 & 56.48 & 66.44 \\
    \rowcolor{tabhighlight}
    Ours & 79.47 & \textbf{61.53} & \textbf{69.36}\\
    \bottomrule
    \end{tabular}
    \end{subtable}
    ~
    \begin{subtable}[t]{.3\textwidth}
    \centering
    \caption{EuroSAT}
    \begin{tabular}{l cc|c}
    \toprule
    & Base & New & Hos \\
    \midrule
    CLIP   & 56.48 & 64.05 & 60.03 \\
    CoOp   & \bf92.19 & 54.74 & 68.69 \\
    CoCoOp & 87.49 & 60.04 & 71.21\\
    ProDA  & 83.90 & 66.00 & 73.88 \\
    \rowcolor{tabhighlight}
    Ours & 92.14 & \bf 73.87 & \textbf{82.00}\\
    \bottomrule
    \end{tabular}
    \end{subtable}
    ~
    \begin{subtable}[t]{.3\textwidth}
    \centering
    \caption{UCF101}
    \begin{tabular}{l cc|c}
    \toprule
    & Base & New & Hos \\
    \midrule
    CLIP   & 70.53 & 77.50 & 73.85 \\
    CoOp   & 84.69 & 56.05 & 67.46 \\
    CoCoOp & 82.33 & 73.45 & 77.64\\
    ProDA  & \textbf{85.23} & 71.97 & 78.04  \\
    \rowcolor{tabhighlight}
    Ours & 84.12 & \textbf{77.74} & \textbf{80.80} \\
    \bottomrule
    \end{tabular}
    \end{subtable}
    }
    \caption{Results (\%) of the \textbf{base-to-new generalization task} on 11 benchmark datasets. We report the accuracy with CLIP ViT-B/16 model on the base classes (Base), the unseen classes (New), and the harmonic mean of both of them (Hos).}
    \label{tab:results_generalization}
    \vspace{-0.4cm}
\end{table*}

\paragraph{CoOp} 
replaces the hand-crafted prompts with automatically generated prompts. Specifically, CoOp introduces $k$ learnable context vectors $\left\{v_1, \ldots, v_k\right\}$ to model the context words of the prompts. 
We define $c_i$ as the word embedding of the $i$-th class name. Then, the prompt of $i$-th class is denoted as
${p}_i=\left\{v_1, \ldots, v_k, c_i\right\}$. 
Therefore, we have the predicted probability of the $i$-th class using CoOp method:
\begin{equation}
P(y=i \mid x)=\frac{\exp \left(\operatorname{cos}\left({{f}_x}, G\left({p}_i\right)\right) / \tau\right)}{\sum_{j=1}^C \exp \left(\operatorname{cos}\left({{f}_x}, G\left({p}_j\right)) / \tau\right)\right.},
\end{equation}
where $G(p_i)$ is the text embedding from text encoder $G$.

\paragraph{CoCoOp} extends CoOp by generating image-conditional prompts. Specifically, CoCoOp uses Meta-Net to generate the residual vector $\pi$ based on each image. Each context token is now obtained by $v_k(x)=v_k+\pi$. The prompt of the $i$-th class $c_i$ is defined as
${p}_i(x)=\left\{v_1(x), \ldots, v_k(x), c_i\right\}$.
As a result, the prediction probability of the $i$-th class is:
\begin{equation}
P(y=i \mid x)=\frac{\exp \left(\operatorname{cos}\left({{f}_x}, G\left({p}_i(x)\right)\right) / \tau\right)}{\sum_{j=1}^C \exp \left(\operatorname{cos}\left({{f}_x}, G\left({p}_j(x)\right) / \tau\right)\right.},
\end{equation}
where $G\left({p}_i(x)\right)$ is the the text embedding conditional on the image $x$ from the text encoder $G$.

\subsection{Our Task-Oriented Mutual Learning Method}
Our method consists of two modules, i.e., \textbf{Class-aware Text Prompts (CTP)} and \textbf{Text-guided Feature Tuning (TFT)}, as shown in Fig. \ref{FIG:3}. 
Compared to CoCoOp, we use CTP to generate class-aware prompts based on task-relevant local image regions instead of the global information. 
Besides, we use CTP to make the image branch directly pay attention to the task-related image region.
We let the two modules be tightly coupled and mutual-beneficial across the training process by optimizing the contrastive loss function.

\paragraph{CTP} learns image conditioned discriminative prompts for finer paying attention to semantic-related regions of the images.
Specifically, in order to obtain the text semantic-related regions of the image, we leverage the prompt $p$ and the image feature ${f}_x$ to calculate the attention matrix $A^t$:
\begin{equation}
A^t=p f_x^T,
\end{equation}
where $A^t\in\mathbb{R}^{M\times N}$ is the image-to-text attention map. $A^t_{i,j}$ represents the similarity between the $i$-th class in the text prompts and the $j$-th patch in the image. 
In this way, we can query the regions of the images that semantically related to the class information by the attention matrix $A^t$. That is,
\begin{equation}
f_x^t=\operatorname{softmax}\left(A^t\right) f_x,
\end{equation}
where $f_x^t$ is the regions correlated to the text of a specific class. 
We use it to obtain augmented class-aware prompts:
\begin{equation}
p^a=p + f_x^t,
\end{equation}
where $p^a$ is the text prompts enhanced by semantically-relevant image regions. 
Let $p^a_i$ be the $i$-th dimension of $p^a$, we then have the predicted probability of the $i$-th class:
\begin{equation}
P(y=i \mid x)=\frac{\exp \left(\operatorname{cos}\left({{f}_x}, G(p^a_i\right) / \tau\right)}{\sum_{j=1}^C \exp \left(\operatorname{cos}\left({{f}_x}, G(p_j^a) / \tau\right)\right.}.
\end{equation}
We generate class-aware prompts instead of fusing identical image semantics with prompts of different classes, bringing category discrimination to the specific downstream tasks.
\vspace{-0.2cm}
\paragraph{TFT} 
leverages text features to guide images to focus on task-related regions. 
Specifically, using the embeddings $g^a$ of the augmented prompts $p^a$ as input, we have attention:
\begin{equation}
A^x=f_x (g^a)^T,
\end{equation}
where $A^x\in\mathbb{R}^{N\times M}$ denotes text-to-image attention map. $A^x_{i,j}$ represents the similarity between the $i$-th patch in the image and the $j$-th class in the text representation. 
Similar to image-to-text, we use it to query the class-related part of the text correlated to the image, augmenting image features:
\begin{equation}
f^a=\operatorname{softmax}(A^x)g^a+{f}_x,
\end{equation}
where $f^a$ is the augmented image embeddings.
We thus let image branch focus on the tasks-related representation. 

\begin{figure*}
	\centering
	\includegraphics[width=0.9\textwidth]{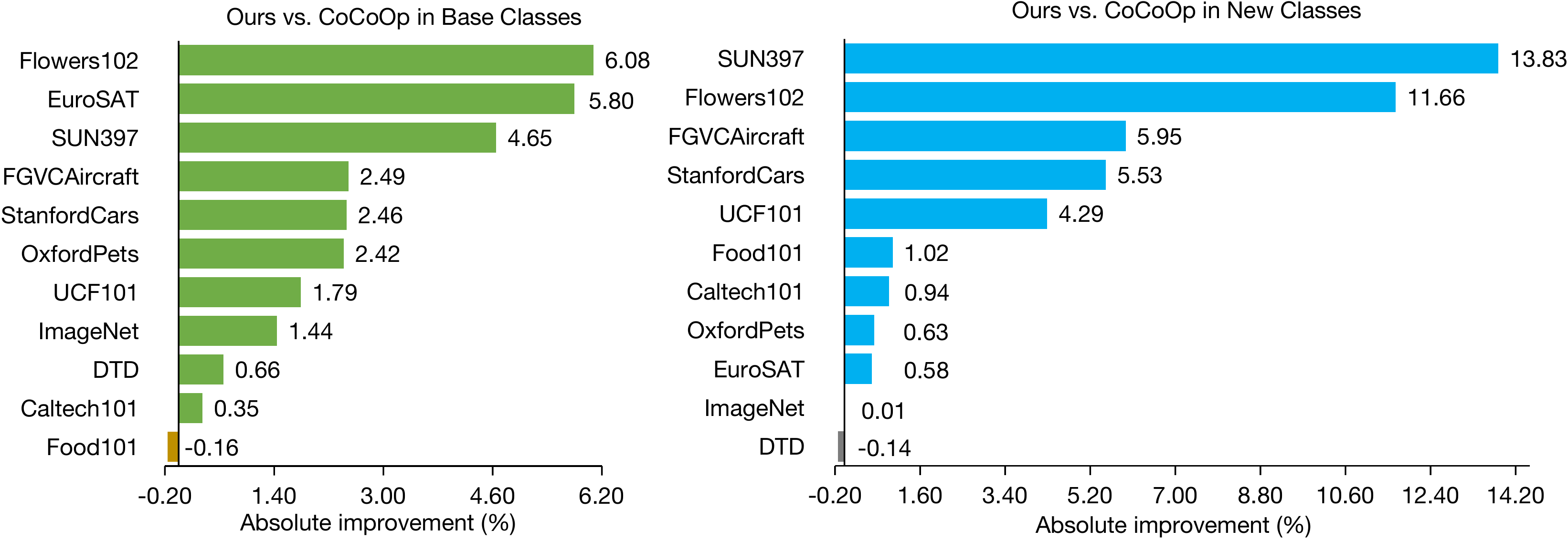}
	\caption{Absolute improvement over CoCoOp in the base-to-new generalization task. Compared to CoCoOp, Our method achieves improvement on both base (left sub-figure) and new (right sub-figure) classes on most of the datasets.
	}
	\label{FIG:absoluteimprovement}
\end{figure*}

\begin{table*}[htp]
\centering
\resizebox{1.0\linewidth}{!}{
 \begin{tabular}{lccccccccccccc}
  \toprule
  
  & 
  \rotatebox{45}{ImageNet} & 
  &
  \rotatebox{45}{Caltech101} & 
  \rotatebox{45}{OxfordPets} & 
  \rotatebox{45}{StanfordCars} & 
  \rotatebox{45}{Flowers102}&
  \rotatebox{45}{Food101}&
  \rotatebox{45}{FGVCAircraft}&
  \rotatebox{45}{SUN397}&
  \rotatebox{45}{DTD}&
  \rotatebox{45}{EuroSAT}&
  \rotatebox{45}{UCF101}&
  \rotatebox{45}{\emph{Average}}
  \\ 
  \midrule
  CLIP&
    68.63&
    &
    89.36&
    88.99&
    65.67&
    70.49&
    89.23&
    27.12&
    65.29&
    46.02&
    54.17&
    69.83&
    66.80\\

    CoOp&
    71.51&
      &
    95.53&
    93.31&
    74.25&
    95.70&
    87.23&
    34.18&
    74.82&
    68.46&
    77.82&
    77.29&
    77.28\\
  CoCoOp&
      71.02 &
      &
      93.43&
    93.93&
    71.21&
    87.34&
    87.39&
    32.03&
    72.32&
    63.84&
    72.78&
    77.40&
    74.79\\
  \rowcolor{tabhighlight}
  \midrule
  Ours&
    \textbf{72.90}&
    &
    \textbf{95.90}&
    \textbf{93.96}&
    \textbf{79.10}&
    \textbf{96.73}&
    \textbf{89.95}&
    \textbf{38.72}&
    \textbf{79.37}&
    \textbf{72.49}&
    \bf 81.00&
    \textbf{83.45}&
    \textbf{80.32}\\ 
  \bottomrule
 \end{tabular}
}
 \caption{Results (\%) of \textbf{16-shot learning task} on 11 datasets.}
 \label{tab2}
 \vspace{-0.2cm}
\end{table*}

\paragraph{Augmented contrastive loss function} is then employed to further align class-aware text embedding and text-guided image features on specific downstream tasks. The predicted probability of the $i$-th class, which is used to calculate the contrastive loss, after mutual augmentation is:
\begin{equation}
P(y=i \mid x)=\frac{\exp \left(\operatorname{cos}\left({f^a}, g^a_i\right) / \tau\right)}{\sum_{j=1}^C \exp \left(\operatorname{cos}\left({f^a}, g_i^a / \tau\right)\right.}.
\end{equation}
Task-targeted semantic information is transferred between the two branches by minimizing the augmented contrastive loss.
We merge probability before and after augmentation:
\begin{equation}
P(y\!=\!i\! \mid \!x)\!=\!\frac{\exp ((\operatorname{cos}\left({f}, g_i\right)\! +\! \lambda\!\left(\operatorname{cos}\left({f^a}, g^a_i\right)) / \tau\right)}{\sum_{j=1}^C \exp ((\operatorname{cos}\left({f}, g_i\right) \!+\! \lambda\!\left(\operatorname{cos}\left({f^a}, g^a_i\right)) / \tau\right)}.
\label{equ:lambda}
\end{equation}
where $\lambda$ is the balance hyper-parameter, which is analyzed in our experiments. 
We let the two different modalities tightly coupled and mutual beneficial across the whole training process by performing the contrastive optimization.

\begin{table*}[htp]
 \centering
\resizebox{1.0\linewidth}{!}{
 \begin{tabular}{l c cccccccccccc}
  \toprule
   &
  \textbf{Source}  & 
  &
  \multicolumn{11}{c}{\textbf{Target}}  \\    
  \cline{2-2} \cline{4-14}  
  & 
  \rotatebox{45}{ImageNet} & 
  &
  \rotatebox{45}{Caltech101} & 
  \rotatebox{45}{OxfordPets} & 
  \rotatebox{45}{StanfordCars} & 
  \rotatebox{45}{Flowers102}&
  \rotatebox{45}{Food101}&
  \rotatebox{45}{FGVCAircraft}&
  \rotatebox{45}{SUN397}&
  \rotatebox{45}{DTD}&
  \rotatebox{45}{EuroSAT}&
  \rotatebox{45}{UCF101}&
  \rotatebox{45}{\emph{Average}}
  \\ 
  \midrule
  CoOp&
  71.51 & &
  93.70 & 
  89.14 & 
  64.51& 
  68.71& 
  85.30& 
  18.47& 
  64.15& 
  41.92& 
  46.39 & 
  66.55& 
  63.88\\ 
  CoCoOp& 
  71.02 & &
  94.43 & 
  90.14 & 
  \textbf{65.32} & 
  \textbf{71.88} & 
  86.06 &
  22.94 &
  \textbf{67.36} &
  45.73 &
  45.37 &
  \textbf{68.21} &
  65.74\\ 
  \midrule
  \rowcolor{tabhighlight}
  Ours& 
  \textbf{72.90} & &
  \textbf{95.73} & 
  \textbf{90.22} & 
  65.14 & 
  69.89 & 
  \textbf{86.38} &
  \textbf{23.32} &
  66.49 &
  \textbf{46.47} &
  \textbf{47.24} &
  67.43 &
  \textbf{66.47}\\ 
  \bottomrule
 \end{tabular}
}
 \caption{Results of \textbf{cross-dataset transfer task}. Each method is trained on the source dataset and evaluated on the target. } 
 \label{cross-dataset}
\end{table*}

\begin{table*}[htp]
 \centering      
 \setlength{\tabcolsep}{6.4mm}{
 \resizebox{1.0\linewidth}{!}{
 \begin{tabular}{lcccccc}
  \toprule
   &
  \textbf{Source}  & 
  &
  \multicolumn{4}{c}{\textbf{Target}}  \\    
  \cmidrule(lr){2-2} \cmidrule(lr){4-7}  
  & 
  ImageNet & 
  &
  ImageNetV2 & 
  ImageNet-Sketch & 
  ImageNet-A & 
  ImageNet-R\\ 
  \midrule
  CLIP& 
  66.73 & 
  &
  60.83 & 
  46.15 & 
  47.77 & 
  73.96 \\ 
  CoOp&
  71.51 & 
  &
  64.20 & 
  47.99 & 
  49.71& 
  75.21\\ 

  CoCoOp& 
  71.02 & 
  &
  64.07 & 
  48.75 & 
  50.63 & 
  76.18\\ 
  \midrule
  \rowcolor{tabhighlight}
  Ours& 
  \textbf{72.90} &
  &
  \textbf{64.57} & 
  \textbf{49.11} & 
  \textbf{50.94} & 
  \textbf{76.68}\\ 
  \bottomrule
 \end{tabular}
}
}
\caption{Results of \textbf{domain generalization task}. Each method is trained on ImageNet and evaluated on ImageNet variants.}
\label{domain-generalization}
\end{table*}

\section{Experiments}
We evaluate the performance of our method on four generalization tasks, including 1) generalization from base classes to new classes; 2) few-shot classification; 3) cross-dataset transfer; 4) domain generalization. After that, we provide extensive ablation studies and in-depth analyses.
\paragraph{Datasets.}
Following \cite{radford2021learning, zhou2022learning}, we use 11 image recognition datasets for the tasks of base-to-new generalization, few-shot classification and cross-dataset transfer.
It contains generic image classification datasets (ImageNet  \cite{deng2009imagenet} and Caltech101  \cite{fei2004learning}), fine-grained classification datasets (Oxford Pets \cite{parkhi2012cats} , StanfordCars  \cite{krause20133d}, Flowers102  \cite{nilsback2008automated}, Food101 \cite{bossard2014food} and FGVCAircraft \cite{maji2013fine}), scene recognition ( SUN397 \cite{xiao2010sun}), action recognition (UCF101  \cite{soomro2012dataset}), texture classification (DTD  \cite{cimpoi2014describing}), and satellite imagery recognition (EuroSAT \cite{helber2019eurosat}). 
For the domain generalization task, we use ImageNet as the source dataset and select ImageNetV2 \cite{recht2019imagenet}, ImageNet-Sketch \cite{wang2019learning}, ImageNet-A \cite{hendrycks2021natural}, and ImageNet-R \cite{hendrycks2021many}, which are the ImageNet variants, as the target.
\paragraph{Training Details.}
By following \cite{zhou2022conditional, zhou2022learning}, we use the best visual backbone available in CLIP, i.e., ViT-B/16, throughout the experiments.  
We train 10 epochs using SGD optimizer with base learning rate of 0.002 and cosine decay schedule.
We set the hyper-parameter $\lambda$ in Eq. (\ref{equ:lambda}) to 0.2 for all experiments, and provide sensitivity analyses in Fig. \ref{FIG:AblationStudy}.
We run all the experiments three times with different random seeds and report the average classification accuracy.

\paragraph{Baselines.}
We compare our method with 4 baselines. (1) Zero-shot CLIP \cite{radford2021learning} with hand-crafted prompts. (2) CoOp \cite{zhou2022learning}, using automatically generated prompts from few data. (3) CoCoOp \cite{zhou2022conditional}, dynamically generating prompts conditioned on the images. (4) ProDA \cite{lu2022prompt}, which learns prompts from few data samples and mitigates the domain gap.

\subsection{Generalization From Base to New Classes}
\label{sec1}
Following the previous works, we split the classes equally into two groups for each dataset: one as base and the other as new. The learnable modules are trained exclusively on the base classes, while evaluation is carried out separately on both the base and new classes to testify generalization ability. We report the results on 11 benchmarks in Table \ref{tab:results_generalization}.
Although compared to CoOp, CoCoOp significantly narrows generalization gap in unseen classed, but it decreases the accuracy in seen classes from 82.69\% to 80.47\%. 
We attribute it to the homogeneous prompts of CoCoOp, which weakens the discriminative semantics of different categories. 
In comparison, our method improves the accuracy in seen classes from 80.47\% to 83.01\% by prompting each text label with corresponding image information. 
Benefit from the mutual learning of our CTP and TFT modules, our method further improves the accuracy in unseen classes from 71.69\% to 75.72 \%, even surpasses the accuracy of CLIP hand-crafted prompts. We provide a detailed comparisons of CoCoOp and our method of per-dataset improvement in Fig. \ref{FIG:absoluteimprovement}. Our method gains significant improvements over CoCoOp in both seen and unseen classes on 10 out of 11 recognition datasets. 
Surprisingly, our method significantly improves CoCoOp by more than 10\% in unseen classes on SUN397 and Flowers102 datasets.
\subsection{Few-Shot Classification}
\label{sec2}
We report few-shot classification results in Table \ref{tab2}. 
Our method surpasses baseline methods on all datasets in the few-shot setting. 
Especially, our method outperforms CoCoOp by 9.39\%, 8.65\%, and 8.22\% on Flowers102, DTD, and EuroSAT, respectively, and the average improvement over 11 datasets is 5.53\%.
Our method also achieve 2\% on the challenging dataset of ImageNet.
The above experiments shows the great discriminative  ability of our method. 

\subsection{Cross-Dataset Transfer}
\label{sec3}
We then evaluate the generalization ability of our method on more challenging cross-dataset tasks.
In this setting, we learn multi-modal prompts on ImageNet of 1000 classes. The effectiveness of the learned prompts is then tested on 10 datasets containing generic and fine-grained image classification, scene recognition, and texture classification.
The results are reported in Table \ref{cross-dataset}.
Our method achieves the best average accuracy on the 11 datasets, especially ImageNet.
It demonstrates the great transfer ability of our method.

\subsection{Domain Generalization}
\label{sec4}
The domain generalization setting evaluates the generalization ability of the model on the target domain that is similar to but different from the source domain. 
Zero-shot CLIP introduces no additional training parameters and exhibits great robustness to naturally distribution shifts. 
Other methods use few samples to train learnable parameters, there is a risk of overfitting the source distribution. Therefore, we conduct experiments using ImageNet as the source domain and evaluate the ability of generalizing to unknown on four ImageNet variants. 
The results are shown in Table \ref{domain-generalization}. Our method achieves significant performance on the 4 ImageNet variant datasets. It verifies that our method improves the classification ability of the source domain dataset while maintaining the generalization on the target domain.

\begin{table*}[htp]
 \scriptsize        
 \setlength{\tabcolsep}{1.4mm}{
\resizebox{1.0\linewidth}{!}{
 \begin{tabular}{llllllllllllllllllll}
  \hline
  \multicolumn{2}{c|}{\multirow{2}{*}{Method}} &
  \multicolumn{3}{c|}{Average}& 
  \multicolumn{3}{c|}{ImageNet}     & 
  \multicolumn{3}{c|}{Caltech101} & 
  \multicolumn{3}{c|}{OxfordPets} & 
  \multicolumn{3}{c|}{StanfordCars} &  
  \multicolumn{3}{c}{Flowers102} \\
  \multicolumn{2}{c|}{}&    
  Base & New  & \multicolumn{1}{c|}{{{Hos}}} & 
  Base & New  & \multicolumn{1}{c|}{{{Hos}}} & 
  Base & New  & \multicolumn{1}{c|}{{{Hos}}} & 
  Base & New  & \multicolumn{1}{c|}{{{Hos}}} & 
  Base & New  & \multicolumn{1}{c|}{{{Hos}}} & 
  Base & New  & {{HOS}}\\ 
  \hline
  \multicolumn{2}{l|}{A}& 
  82.69 &  63.22   & \multicolumn{1}{c|}{71.66} &         
  76.47 &67.88 & \multicolumn{1}{c|}{71.92}  & 
  98.00 & 89.81 &   \multicolumn{1}{c|}{93.73}     & 
  93.67 & 95.29 & \multicolumn{1}{c|}{94.47} & 
  \bf 78.12 & 60.40& \multicolumn{1}{c|}{68.13} & 
  \bf 97.60 & 59.67  & 74.06 \\
  \multicolumn{2}{l|}{B}& 
  82.38 & 72.44 & \multicolumn{1}{c|}{76.64} & 
  76.96 & 69.62 & \multicolumn{1}{c|}{73.11} & 
  98.23 & 94.24 & \multicolumn{1}{c|}{96.19} & 
  95.61 & \bf97.97 & \multicolumn{1}{c|}{\bf96.78} & 
  74.62 & 73.68 & \multicolumn{1}{c|}{74.15} &
  96.72  & 66.42 & 78.76  \\
  \multicolumn{2}{l|}{C}&    
  82.93 & 72.98  & \multicolumn{1}{c|}{77.24} & 
  77.21  & 69.86 & \multicolumn{1}{c|}{73.35}  & 
  \bf98.44 & 92.72  & \multicolumn{1}{c|}{95.49} & 
  95.49 & 97.81 & \multicolumn{1}{c|}{96.64}  & 
  75.84 & \bf74.53 & \multicolumn{1}{c|}{75.18}  & 
  97.32 & 74.86  & 84.63   \\
  \multicolumn{2}{l|}{Ours}& 
  \bf83.01 & \bf75.72 & \multicolumn{1}{c|}{\bf79.02}  & 
  \bf77.42 & \bf70.44 & \multicolumn{1}{c|}{\bf73.77} & 
  98.31 & \bf94.75 & \multicolumn{1}{c|}{\bf96.50} & 
  \bf95.86 & 97.55 & \multicolumn{1}{c|}{96.70} & 
  76.29 & 74.17 & \multicolumn{1}{c|}{\bf75.22} & 
  97.36 & \bf77.70 & \bf86.43 \\
  \hline\hline                   
  \multicolumn{2}{c|}{\multirow{2}{*}{Method}} &
  \multicolumn{3}{c|}{Food101}& 
  \multicolumn{3}{c|}{FGVCAircraft} & 
  \multicolumn{3}{c|}{SUN397} & 
  \multicolumn{3}{c|}{DTD} & 
  \multicolumn{3}{c|}{EuroSAT} & 
  \multicolumn{3}{c}{UCF101} \\
  \multicolumn{2}{l|}{}& 
  Base  & New & \multicolumn{1}{c|}{{{Hos}}} & 
  Base  & New & \multicolumn{1}{c|}{{ {Hos}}} & 
  Base  & New & \multicolumn{1}{c|}{{ {Hos}}} & 
  Base  & New & \multicolumn{1}{c|}{{ {Hos}}} & 
  Base  & New & \multicolumn{1}{c|}{{ {Hos}}} & 
  Base  & New & \multicolumn{1}{c}{{ {Hos}}} \\ 
  \hline
  \multicolumn{2}{l|}{A}& 
  88.33 &82.26 & \multicolumn{1}{c|}{85.19} &   
  \bf 40.44 & 22.30 & \multicolumn{1}{c|}{28.75} &
  80.60 &65.89 & \multicolumn{1}{c|}{72.51} &  
  79.44 &41.18& \multicolumn{1}{c|}{54.24} & 
  \bf 92.19 &  54.74 & \multicolumn{1}{c|}{68.69} & 
  84.69 &56.05 & 67.46    \\
  \multicolumn{2}{l|}{B}&
  90.30 & 91.47 & \multicolumn{1}{c|}{90.88} & 
  36.41 & 34.39 & \multicolumn{1}{c|}{35.37} & 
  81.73 & 76.89 & \multicolumn{1}{c|}{79.24} & 
  80.18 & 51.79 & \multicolumn{1}{c|}{62.93} & 
  91.70 & 67.62 & \multicolumn{1}{c|}{77.84} &  
  83.72 & 72.72  & 77.83  \\
  \multicolumn{2}{l|}{C}&
  \bf90.56 & 91.65 & \multicolumn{1}{c|}{91.10} & 
  37.82 & 33.17 & \multicolumn{1}{c|}{35.34} & 
  \bf82.29 & 76.24 & \multicolumn{1}{c|}{79.15} & 
  \bf81.71 & 54.74 & \multicolumn{1}{c|}{65.56} & 
  90.10 & 60.52 & \multicolumn{1}{c|}{72.41} &  
  \bf85.40 &76.68 & \bf80.81         \\
  \multicolumn{2}{l|}{Ours}&
  90.54 & \bf92.31 & \multicolumn{1}{c|}{\bf91.42} & 
  39.49 & \bf35.37 & \multicolumn{1}{c|}{\bf37.32} & 
  82.16 & \bf77.49 & \multicolumn{1}{c|}{\bf79.76} & 
  79.47 & \bf61.53 & \multicolumn{1}{c|}{\bf69.36} &  
  92.14 & \bf73.87 & \multicolumn{1}{c|}{\bf82.00} & 
  84.12 & \bf77.74 & 80.80 \\
  \hline
 \end{tabular}
}
}
\caption{\textbf{Ablation studies} of our method on 11 datasets. Three ablation cases are considered: \textbf{A}: Ours w\slash{}o TFT w\slash{}o CTP. \textbf{B}: Ours w\slash{}o TFT. \textbf{C}: Ours w\slash{}o CTP.
 TFT is the Text-guided Feature Tuning, and CTP is the Class-aware Text Prompts.  
 }
\label{ablation}
\end{table*}

\subsection{Ablation Analysis}
\paragraph{Effectiveness of each module.}
To evaluate the effectiveness of Class-aware Text Prompts (CTP) and Text-guided Feature Tuning (TFT) of our method, we conduct ablation experiments on 11 datasets, as reported in Table \ref{ablation}. 
In most cases, each module significantly improves the performance of the model. 
For average results, CTP and TFT improves the results by 5.58\% and 4.98\%, respectively, and the combination of them improves the results by 7.36\%. 
It show the effectiveness of the two branches of text-to-image and image-to-text, the mutual learning of the two modules further improves the performance on downstream tasks.

\paragraph{Comparison of different structure design of multi-modal mutual learning.}
To further provide in-depth analysis about our mutual learning, we further explore two vanilla structures: 
(1) MLP-PL: The image features are forwarded to a block of Linear-ReLU-Linear, borrowed from \cite{zhou2022conditional}, and then added to the text for augmenting it (the same to us). 
(2) MLP-FT: The text prompts are forwarded to the Linear-ReLU-Linear block, and then added to the image for augmenting it (the same to us). 
In comparison, our class-aware text prompts (CTP) module and text-guided feature tuning (TFT) module, adopt text-image attention to learn the augmented features instead of the Linear-ReLU-Linear block. 
We report the results of the different designs in Table \ref{tab:ab:component}.
First, we find that combining MLP-PL \& MLP-FT and CTP \& TFT can both improve the results compared with using either of them. 
It indicates that both prompt learning and feature tuning are important to achieve better results. 
Second, compared with the design of Linear-ReLU-Linear block, our design of text-image attention further improves performance by 0.81\% and 1.3\% for prompt learning and feature tuning, respectively. 
It demonstrates the effectiveness of our design of attention, which helps the model to focus on class-aware and task-related semantics. 
Third, compared with CoOp, both of the designs could improve the final results by large margins. The key factor of our mutual learning to achieve significant performance is the task-related alignment of vision and language in latent space.

\begin{figure}[htbp]
    \centering
	\begin{subfigure}[b]{0.47\linewidth}
    {
      \centering
	   \includegraphics[width=1.0\textwidth]{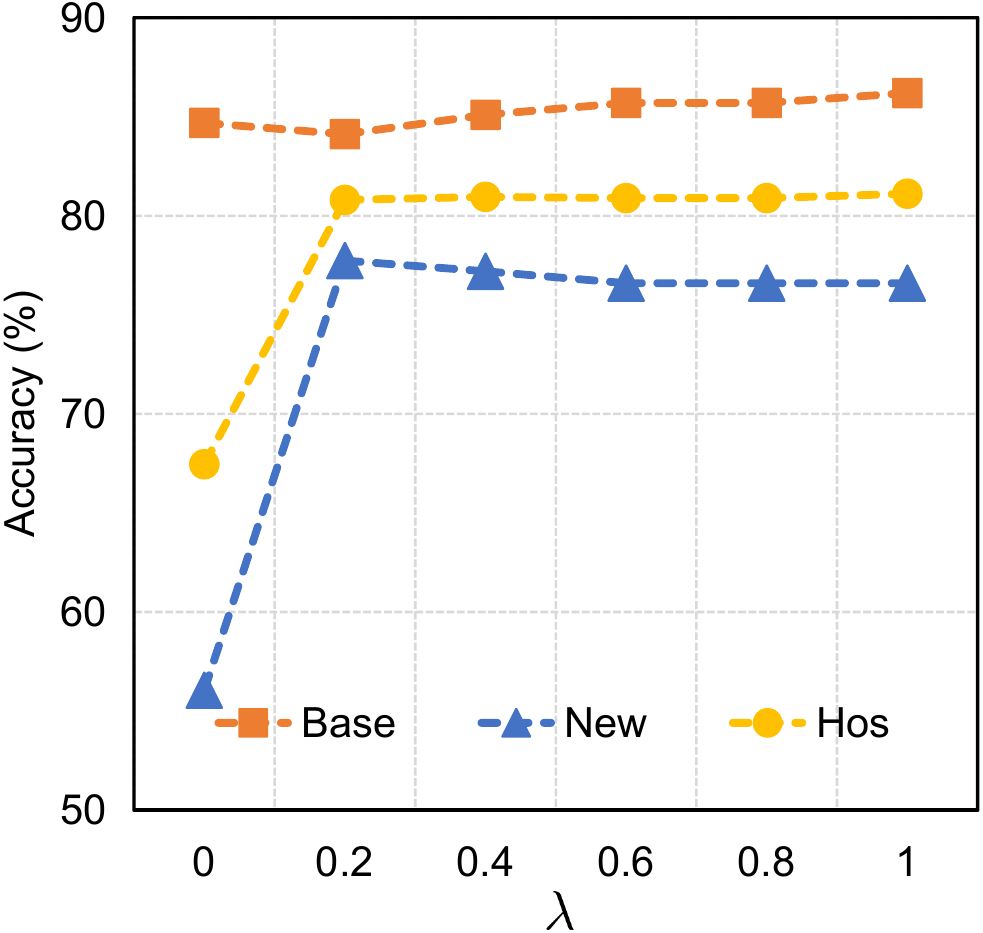} 
	}
    \end{subfigure}
    \hfill
	\begin{subfigure}[b]{0.47\linewidth}
    {
      \centering
	   \includegraphics[width=1.0\textwidth]{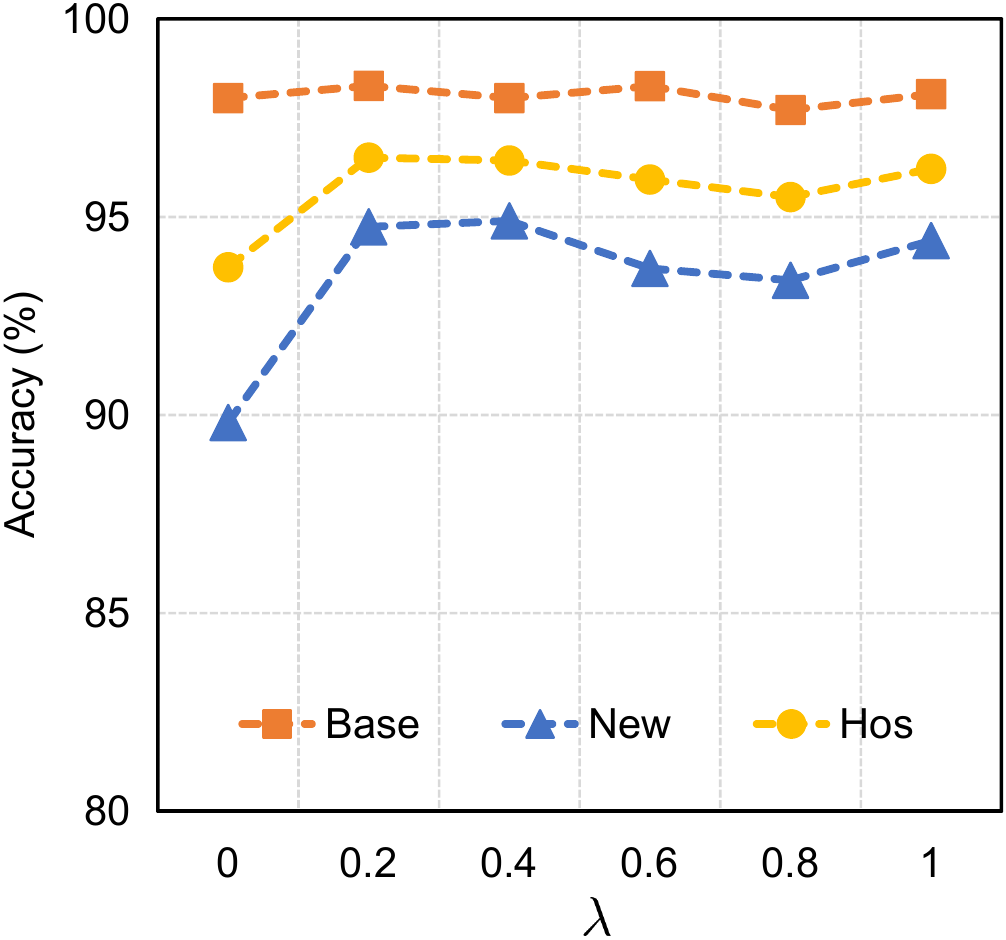} 
	}
    \end{subfigure}
	\caption{\textbf{Sensitivity analysis of ${\lambda}$}, with base, new, and hos metrics, on UCF101 (left) and Caltech101 (right) datasets.}
	\label{FIG:AblationStudy}
 \vspace{-0.2cm} 
\end{figure}

\paragraph{Sensitivity Analysis of $\lambda$.}
We evaluate the parameter sensitivity of $\lambda$ of Eq. (\ref{equ:lambda}) in Fig. \ref{FIG:AblationStudy}.
 The results suggest that the performance of our method is generally robust to $\lambda$, indicating a wide range of $\lambda$ works well in downstream tasks.

\begin{table}
  \centering
  \setlength{\tabcolsep}{5.4mm}{
  \resizebox{1.0\linewidth}{!}{
  \begin{tabular}{c|c|c|c|c}
    \toprule
    \multicolumn{2}{c|}{Prompt Learning} & 
    \multicolumn{2}{c|}{Feature Tuning} & \multirow{2}{*}{ Accuracy (\%)} \\
    \cmidrule(lr){1-4}
    MLP-PL & CTP &  MLP-FT & TFT \\
    \midrule
    &         &   &  & 71.66 {({CoOp})} \\
    \checkmark &  &  &  & 75.83 {(\hgreen{+4.17})}\\
    & \checkmark &  &   &  76.64 {(\hgreen{+4.98})}\\
    &  & \checkmark & &  75.94 {(\hgreen{+4.28})}\\
    &  &  & \checkmark &  77.24 {(\hgreen{+5.58})}\\
    \checkmark &  & \checkmark & &  77.05 {(\hgreen{+5.39})}\\
    \midrule
    & \checkmark &  & \checkmark & { {\bf79.02} {(\hgreen{+7.36})}}\\
    \bottomrule
  \end{tabular}
  }
  }
\caption{Comparison of different structures for prompt learning and feature tuning. The average results of harmonic mean of from-base-to-new generalization task on 11 datasets are reported. 
In compared to our attention design in CTP and TFT modules, MLP-PL and MLP-FT are designed using the Linear-ReLU-Linear block setting of \cite{zhou2022conditional}. 
Improvements over the baseline of CoOp, are marked in \hgreen{green}.}
  \label{tab:ab:component}
\end{table}

\section{Conclusion}
In this paper, we introduce task-oriented multi-modal mutual learning for adapting large vision-language models to downstream vision tasks. 
We propose class-aware text prompt and text-guided feature tuning to unleash the potential of the vision-language model by re-activating its task-related representation abilities. 
Our method yields impressive generalization performance on a wide range of vision tasks and datasets. 
We hope the presented findings and insights in this paper could benefit the following works in designing more efficient and effective adaptation methods. 
For the future work, we think it is interesting to extend the adaptation of vision language models to more vision tasks, such as semantic segmentation, object detection, etc. 


\clearpage
{\small
\bibliographystyle{ieee_fullname}
\bibliography{egbib}
}



\end{document}